\documentclass{article} 
\usepackage{iclr2027_conference,times}

\usepackage{amsmath,amsfonts,bm}

\def\eqref#1{equation~\ref{#1}}

\def\1{\bm{1}}

\DeclareMathAlphabet{\mathsfit}{\encodingdefault}{\sfdefault}{m}{sl}
\SetMathAlphabet{\mathsfit}{bold}{\encodingdefault}{\sfdefault}{bx}{n}

\usepackage{hyperref}
\usepackage{url}
\usepackage{graphicx}
\usepackage{wrapfig}
\usepackage{booktabs}
\usepackage{hyperref}
\usepackage{xcolor}

\title{
\textsc{Valerant}: An Automatic Na\underline{V}ig\underline{A}b\underline{L}\underline{E} Game Map Generato\underline{R} via \underline{A}ction-Co\underline{N}ditioned World Model Explora\underline{T}ion
}

\author{
    Yiran Qiao$^{1}$,
    Feng Wang$^{2}$,
    Jing Ma$^{1}$\thanks{\quad Corresponding author.}
    \\
    $^{1}$Case Western Reserve University, $^{2}$Johns Hopkins University \\
    \texttt{\{yxq350,jxm1384\}@case.edu}, \texttt{wangf3014@gmail.com}
}

\iclrfinalcopy 
\begin{document}

\maketitle
\fancyhead{}

\begin{abstract}
    World Action Models (WAMs) couple predictive world modeling with action generation, allowing anticipated future states to guide agent behavior. Although WAMs are rapidly advancing embodied AI, general-purpose counterparts remain largely unexplored in games. Existing game-oriented approaches often combine action-conditioned world models with external policies and reward functions to realize WAM-like decision-making, yet they operate mainly in 2D visual observation space and do not instantiate persistent 3D geometry. Extending this paradigm to 3D games introduces a distinct challenge. In autonomous driving and robotics, the physical environment exists independently of the model, providing a persistent 3D world in which selected actions can be executed. Games have no such external substrate; the virtual world itself must be instantiated. Most playable games require a persistent and navigable space, while 3D games additionally require explicit geometry that supports movement and interaction. Action-conditioned video rollouts provide visual observations but not this spatial representation. We present \textsc{Valerant}, a training-free framework that transforms a pretrained action-conditioned world model into a WAM for exploring and constructing 3D game maps. By coupling predictive visual rollouts with SLAM-based spatial reconstruction and exploration-driven action selection, \textsc{Valerant} progressively transforms a single image into a persistent 3D game map. This framework extends WAM-based interaction beyond 2D visual simulation and offers a new approach to reducing manual effort in 3D game-map creation. \href{https://yrqiao.github.io/VALERANT/}{\textcolor{blue}{[Project Page]}}
\end{abstract}
\section{Introduction}
Rapid advances in large-scale generative modeling have established content synthesis as one of the most active areas of modern artificial intelligence, giving rise to the broad paradigm of AI-generated content (AIGC) \citep{peebles2023scalable,podell2024sdxl,wu2025vila,wan2025wan}. Modern generative models can synthesize high-quality images, videos, and multimodal sequences under increasingly flexible forms of conditioning. Among their dominant technical foundations, diffusion models generate data by learning to reverse a progressive corruption process \citep{ho2020denoising,rombach2022high}, whereas flow-matching models learn a continuous velocity field that transports samples from a simple prior distribution to the data distribution \citep{lipman2022flow,liu2022flow}. Building on increasingly coherent video generation, recent work has begun to formulate video generators as observation-space world models (WMs) \citep{brooks2024video,wang2024driving,alhaija2025cosmos,qin2024worldsimbench}. By predicting temporally coherent visual trajectories, these models capture regularities in object motion, scene evolution, and the observable outcomes of physical interactions, thereby approximating how a world evolves without requiring an explicit representation of its underlying 3D state. When actions are introduced as conditioning signals, the resulting action-conditioned world models predict future observations under a prescribed action or action sequence \citep{yang2023learning,valevski2025diffusion,che2025gamegen,bar2025navigation,parkerholder2024genie2,genie3}. Crucially, predicting the consequence of a given action does not by itself determine which action should be taken. World Action Models (WAMs) take a further step by coupling predictive world modeling with action generation or selection \citep{wang2026world,zhang2026world}. This coupling can be realized through a cascaded pipeline that derives actions from predicted futures or through a unified model that jointly models future states and actions \citep{du2023learning,zhao2025cot}. Rather than merely simulating what may happen under an externally specified action, WAMs use anticipated world evolution to inform what an agent should do. Taken together, this progression extends generative modeling from passive content synthesis, through controllable prediction, to predictive decision-making in dynamic environments.

The practical value of this shift is particularly evident in autonomous driving and robotics, where WAMs use predicted world evolution to guide motion planning and policy generation \citep{xia2026drivelaw,zhao2025cot,kim2026cosmos, xiong2026unidrive}. The motivation is straightforward: collecting closed-loop experience and evaluating policies in the physical world are expensive, time-consuming, and difficult to scale, while failed trials may damage hardware or create unacceptable safety risks \citep{li2024evaluating,quevedo2026worldgym}. By coupling prospective world prediction with action generation, WAMs allow possible behaviors and their consequences to inform decisions before costly physical deployment. Game creation presents a fundamentally different setting. General-purpose WAMs remain largely unexplored in games, particularly for applications that require the underlying 3D world itself to be instantiated. Existing approaches instead follow two related routes: model-based agents combine learned WMs with external policies or reward functions to optimize behavior, while interactive game WMs generate action-conditioned visual observations from user inputs \citep{alonso2024diffusion,hafner2025mastering,valevski2025diffusion,li2025hunyuan,wang2026matrix}. Although these systems can realize WAM-like interaction or decision-making at the pipeline level, their learned environments remain represented by 2D frames or latent observation sequences rather than persistent, explicit 3D geometry. Comparable modular WM--policy systems are also used in autonomous driving and robotics \citep{wang2024driving,bar2025navigation,gao2026dreamdojo}. Importantly, however, the worlds modeled in these physical domains already exist independently of the learned system. A vehicle or robot operates in a persistent physical 3D environment, which provides the spatial substrate against which model predictions can be grounded and from which new observations can be obtained after execution. In game creation, by contrast, the world being modeled is also the artifact that must be generated. There is no independently existing physical environment that can provide a persistent 3D scene behind the synthesized observations. Most first-person shooters require a persistent, navigable 3D space in which camera motion, collision, visibility, and object interactions can be resolved consistently over time. Observation-space WMs and existing WM--policy systems therefore cannot directly provide the explicit 3D maps required for constructing such games. Bridging WAM-based exploration with explicit 3D map construction is thus necessary to extend these models from interactive visual simulation to practical game creation.

To address this limitation, we introduce \textsc{Valerant}, an automatic game-map generator that transforms a single input image into a navigable 3D environment. Without additional training, \textsc{Valerant} converts a pretrained action-conditioned WM into a WAM by coupling its predictive rollouts with exploration-driven action selection. \textsc{Valerant} instantiates a virtual agent that queries the WM with a finite set of candidate actions at each exploration step, producing counterfactual video rollouts from a common pre-action state. Each rollout is processed by visual SLAM \citep{teed2021droid,matsuki2024gaussian,murai2025mast3r} to estimate the camera trajectory and reconstruct the observed 3D structure. An exploration policy scores the resulting candidates according to their mapping utility, after which the agent returns to the pre-action state and executes only the highest-scoring action. The selected observations are integrated into a persistent map, and this process is repeated until the exploration budget is exhausted, yielding a consolidated 3D point-cloud game map. Our contributions are threefold:
\begin{itemize}
    \item We introduce \textsc{Valerant}, a unified training-free framework that converts a pretrained action-conditioned WM into a WAM and couples its exploration with SLAM-based reconstruction for automatic game-map generation.
    \item We develop a counterfactual exploration procedure in which alternative WM rollouts guide action selection and are converted into a persistent 3D environment, bridging the gap between visual world simulation and the explicit spatial instantiation required by games.
    \item We establish a new direction for game-map authoring in which a single visual concept can be expanded into navigable 3D geometry, thereby reducing game developers' reliance on labor-intensive and cumbersome map-construction workflows.
\end{itemize}
\section{Related Work}

\subsection{World Model and World Action Model}

\looseness -1
World models learn predictive representations of environment dynamics to support planning and control. Early approaches compressed high-dimensional observations into latent states and learned action-conditioned transitions within these compact spaces \citep{ha2018world,hafner2019learning,hafner2019dream}. Subsequent methods improved the scalability, robustness, and generality of latent world modeling, enabling agents to learn across increasingly diverse control tasks through imagined trajectories \citep{hansen2024td,hafner2025mastering}. These models are primarily optimized to capture task-relevant dynamics for decision-making, rather than to synthesize photorealistic and spatially persistent environments. 
Advances in video generation have shifted world modeling toward the direct prediction of visual futures. iVideoGPT \citep{wu2024ivideogpt} represents observations, actions, and rewards as interleaved tokens for scalable interactive prediction, while AVID \citep{rigter2024avid} adapts pretrained video diffusion models into action-conditioned simulators using lightweight domain-specific adapters. In robotics, IRASim \citep{zhu2025irasim} generates fine-grained robot--object interactions conditioned on action trajectories. These systems are action-conditioned WMs: actions are supplied as conditions, and the models predict their observational consequences rather than generating the actions themselves. 

Following recent formalizations \citep{wang2026world,zhang2026world}, we reserve the term World Action Model (WAM) for systems that couple forward world prediction with action generation or selection. This coupling can be implemented through cascaded architectures that derive actions from predicted futures or joint architectures that model future states and actions within a unified framework \citep{du2023learning,zhao2025cot,bi2026motus,kim2026cosmos}. Games provide a natural testbed for these models because both observations and actions can be collected at scale. WHAM \citep{kanervisto2025world} jointly models gameplay visuals and controller actions, representing a game-specific joint WAM for gameplay ideation. By contrast, Oasis \citep{decart2024oasis}, MineWorld \citep{guo2025mineworld}, and Matrix-Game \citep{wang2026matrix} are action-conditioned WMs that generate future gameplay observations from user controls. Recent work has further extended action-conditioned generation to environments containing reactive agents and complex multi-agent behavior \citep{agarwal2026combat}. Despite their increasing visual fidelity and controllability, these methods primarily realize their generated worlds as autoregressive streams of visual observations. Geometry-aware WMs have begun to address spatial forgetting by introducing persistent point-cloud memory or latent 3D scene states \citep{wu2026video,garcin2026beyond}. Their primary objective, however, is to improve the spatial consistency and long-horizon stability of visual generation rather than to construct a reusable game map. In contrast, \textsc{Valerant} starts from a pretrained action-conditioned WM and, without additional training, couples its counterfactual rollouts to an exploration policy, thereby forming a cascaded WAM whose selected observations are reconstructed into an explicit 3D point-cloud game map.

\subsection{Visual SLAM and Online Mapping}
Simultaneous localization and mapping (SLAM) jointly estimates an agent's trajectory and incrementally constructs a spatial representation of its surroundings from sequential observations. Classical visual SLAM systems rely on geometric correspondences, bundle adjustment, and loop closure to maintain globally consistent camera poses and maps, as exemplified by ORB-SLAM3 \citep{campos2021orb}. Learning-based approaches have since incorporated learned components into this pipeline: DROID-SLAM \citep{teed2021droid} and iSLAM \citep{fu2024islam} perform recurrent optimization over camera poses and dense depth, while recent systems such as MASt3R-SLAM \citep{murai2025mast3r}, SLAM3R \citep{liu2025slam3r}, and VGGT-SLAM \citep{maggio2026vggt} exploit learned 3D reconstruction priors to recover dense geometry from monocular image streams. SLAM has also been tightly coupled with embodied exploration, where an agent repeatedly selects an action, moves through its environment, acquires new observations, and updates its map online \citep{chaplot2020learning}. Consequently, both robotic SLAM and \textsc{Valerant} follow an incremental mapping process in which exploration continually reveals new parts of the environment. Their central distinction lies instead in the source of these observations. Conventional embodied systems obtain sensor measurements by moving through an existing real or simulated 3D environment. In \textsc{Valerant}, no such environment is initially available. A frozen action-conditioned video WM generates the predicted visual consequences of externally specified candidate actions, while an exploration policy evaluates the resulting what-if trajectories and selects the action to execute. SLAM then estimates the corresponding camera motion and integrates the selected observations, progressively transforming transient WM rollouts into a persistent and navigable 3D point-cloud map.

\subsection{Counterfactual Reasoning and What-If Rollouts}


\looseness -1
Counterfactual reasoning asks how the future would change under an alternative action while the preceding context remains fixed \citep{rubin1974estimating,holland1986statistics,pearl2003causality}. In the physical world, however, only the outcome of the executed action can be observed; the mutually exclusive futures associated with unchosen actions remain inaccessible. Learned world models provide a practical means of exploring these alternatives by simulating candidate actions before execution, an idea widely used in model-based planning \citep{hansen2024td}. Action-conditioned visual WMs further render these alternatives as predicted video trajectories, supporting the comparison of possible maneuvers in embodied navigation and autonomous driving \citep{yu2022modular,wang2024driving,bar2025navigation}. Such WMs are naturally suited to counterfactual-style reasoning because they can preserve a common observation history while varying the candidate action. This predictive capability becomes part of a WAM when the resulting futures are coupled to action generation or selection. \textsc{Valerant} realizes this coupling without additional training: it evaluates alternative WM rollouts according to their mapping utility and selects the action that most effectively reveals unobserved 3D space.
\begin{figure}
    \centering
    \includegraphics[width=1\linewidth]{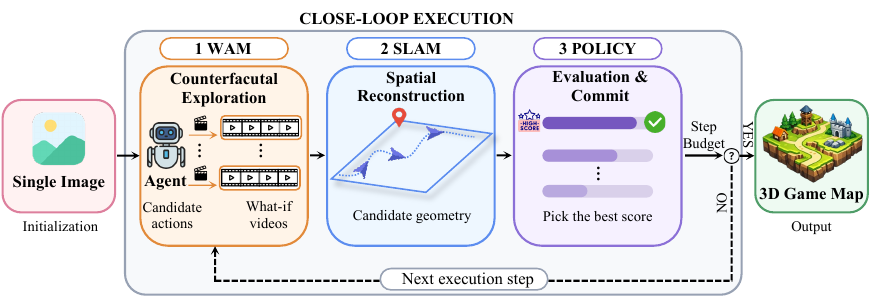}
    \caption{Overview of \textsc{Valerant}. Starting from a single input image, a virtual agent queries the WAM with candidate actions to generate alternative what-if video rollouts. Visual SLAM reconstructs the corresponding 3D geometry, and an exploration policy selects the branch with the highest mapping utility. The selected observations are committed to a persistent point-cloud map, which is iteratively expanded into the final 3D game map.}
    \label{fig1}
\end{figure}
\section{Method}

\subsection{Preliminaries}

\paragraph{World Models.}
A world model (WM) approximates the transition dynamics of an environment by predicting how its observations evolve under externally specified actions \citep{wang2026world}. Let $\mathbf{o}_t$ denote the visual observation at environment step $t$. We define the visual context as $\mathbf{O}^{\mathrm{c}}_t=(\mathbf{o}_{t-K+1},\ldots,\mathbf{o}_t)$, a candidate action sequence as $\mathbf{A}_t=(\mathbf{a}_t,\ldots,\mathbf{a}_{t+H-1})$, and the corresponding future observations as $\mathbf{O}^{\mathrm{f}}_t=(\mathbf{o}_{t+1},\ldots,\mathbf{o}_{t+H})$, where $K$ and $H$ denote the context length and prediction horizon, respectively. An observation-space WM parameterized by $\theta$ generates a possible future rollout according to
\begin{equation}
    {\mathbf{O}}^{\mathrm{f}}_t
    \sim
    p_{\theta}\!\left(
        \,\cdot\,
        \mid
        \mathbf{O}^{\mathrm{c}}_t,
        \mathbf{A}_t
    \right),
    \label{eq:world_model}
\end{equation}
where $p_{\theta}$ denotes the conditional distribution over future visual observations learned by the WM. Modeling a distribution allows the WM to represent multiple plausible futures that may follow the same observation history and action sequence. Importantly, the action sequence is provided as a condition: the WM predicts its possible visual consequences but does not determine which action should be executed.

\paragraph{Action-Conditioned Video Diffusion Models.}
An action-conditioned video diffusion WM instantiates the conditional distribution in \eqref{eq:world_model} with a latent video generator \citep{wang2026matrix}. Let $\mathcal{E}$ and $\mathcal{D}$ denote a video encoder and decoder, respectively. We encode the visual context as $\mathbf{h}_t=\mathcal{E}(\mathbf{O}^{\mathrm{c}}_t)$ and the target future video as $\mathbf{z}_0=\mathcal{E}(\mathbf{O}^{\mathrm{f}}_t)$. Following the flow-matching (FM) formulation, a noisy latent at diffusion time $\tau\in[0,1]$ is constructed as
\begin{equation}
    \mathbf{z}_{\tau}
    =
    (1-\tau)\mathbf{z}_0
    +
    \tau\boldsymbol{\epsilon},
    \qquad
    \boldsymbol{\epsilon}\sim\mathcal{N}(\mathbf{0},\mathbf{I}),
    \label{eq:latent_interpolation}
\end{equation}
and the conditional velocity field $v_{\theta}$ is learned using
\begin{equation}
    \mathcal{L}_{\mathrm{FM}}
    =
    \mathbb{E}\!\left[
        \left\|
        v_{\theta}\!\left(
            \mathbf{z}_{\tau},
            \tau
            \mid
            \mathbf{h}_t,
            \mathbf{A}_t
        \right)
        -
        \left(\boldsymbol{\epsilon}-\mathbf{z}_0\right)
        \right\|_2^2
    \right].
    \label{eq:flow_matching}
\end{equation}
At inference time, the learned flow is integrated from Gaussian noise toward the video latent and decoded by $\mathcal{D}$ to obtain ${\mathbf{O}}^{\mathrm{f}}_t$. Consequently, different candidate action sequences conditioned on the same visual context produce alternative future video rollouts. In \textsc{Valerant}, the video WM remains frozen and is used only for action-conditioned rollout generation.


\paragraph{Visual Simultaneous Localization and Mapping.}
Visual simultaneous localization and mapping (SLAM) jointly estimates camera motion and reconstructs 3D scene structure from visual observations \citep{teed2021droid,murai2025mast3r}. Given an observation sequence $\mathbf{O}_{1:L}=(\mathbf{o}_1,\ldots,\mathbf{o}_L)$, visual SLAM estimates the camera trajectory $\mathbf{T}_{1:L}$, where $\mathbf{T}_i\in\mathrm{SE}(3)$ is the camera pose at frame $i$, together with a spatial map $\mathcal{M}$. This process can be expressed abstractly as
\begin{equation}
    \left(
        \hat{\mathbf{T}}_{1:L},
        \hat{\mathcal{M}}
    \right)
    =
    \underset{\mathbf{T}_{1:L},\,\mathcal{M}}{\arg\min}\;
    \mathcal{J}_{\mathrm{SLAM}}\!\left(
        \mathbf{O}_{1:L};
        \mathbf{T}_{1:L},
        \mathcal{M}
    \right),
    \label{eq:visual_slam}
\end{equation}
where $\mathcal{J}_{\mathrm{SLAM}}$ measures the visual and geometric consistency between the observations, camera poses, and reconstructed scene. In \textsc{Valerant}, each generated rollout $\hat{\mathbf{O}}^{\mathrm{f}}_{t,m}$ is processed by visual SLAM to recover its camera trajectory and register its observed geometry in a common 3D coordinate system.

\subsection{Overview}
\label{sec:method_overview}

Given a single input image $\mathbf{o}_0$, \textsc{Valerant} progressively explores the synthesized environment and constructs a navigable 3D game map. At exploration step $t$, the system maintains a visual context $\mathbf{O}^{\mathrm{c}}_t$, a current camera pose $\mathbf{T}_t$, and a global map $\mathcal{M}_t$. The pretrained video WM remains frozen throughout this process, while visual SLAM and the exploration policy are applied entirely at inference time. By coupling the WM with geometric reconstruction and action selection, \textsc{Valerant} turns its passive visual predictions into a closed-loop exploration process without additional training.

At each step, the system checkpoints the current WM and SLAM states and evaluates five candidate action sequences:
\begin{equation}
    \mathcal{A}_t
    =
    \left\{
        \mathbf{A}^{(1)}_t,
        \ldots,
        \mathbf{A}^{(5)}_t
    \right\}
    =
    \left\{
        \mathtt{uw},
        \mathtt{jw},
        \mathtt{lw},
        \mathtt{jq},
        \mathtt{lq}
    \right\}.
    \label{eq:candidate_actions}
\end{equation}
Here, $\mathtt{uw}$ denotes moving straight forward without turning, while $\mathtt{jw}$ and $\mathtt{lw}$ denote turning left and right, respectively, followed by forward movement. The remaining actions, $\mathtt{jq}$ and $\mathtt{lq}$, denote in-place rotations to the left and right. The resulting action space therefore contains three forward-moving candidates and two rotation-only candidates.

Starting from the same pre-action context $\mathbf{O}^{\mathrm{c}}_t$, the WM generates a separate future rollout $\mathbf{O}^{\mathrm{f}}_{t,m}$ for each candidate $\mathbf{A}^{(m)}_t$ following \eqref{eq:world_model}. Because all candidates branch from an identical visual context, their generated rollouts provide directly comparable what-if outcomes of the available actions. Rather than estimating their consequences using an additional learned predictor, \textsc{Valerant} explicitly generates and reconstructs every candidate future.

Visual SLAM processes each generated rollout $\mathbf{O}^{\mathrm{f}}_{t,m}$ according to \eqref{eq:visual_slam}, producing an estimated camera trajectory $\hat{\mathbf{T}}_{t,m}$ and candidate scene geometry $\hat{\mathcal{M}}_{t,m}$. Each candidate is therefore evaluated using its reconstructed motion and observed 3D structure rather than its visual appearance alone. We define its mapping utility using exploration coverage $\mathcal{C}_{t,m}$ and forward progress $\mathcal{P}_{t,m}$, while $\mathcal{B}_{t,m}\in\{0,1\}$ indicates whether the candidate violates the collision-free constraint. The selected branch is
\begin{equation}
    m^{\star}
    =
    \underset{
        m\in\{1,\ldots,5\}:
        \mathcal{B}_{t,m}=0
    }{\arg\max}
    \left(
        \lambda_{\mathrm{cov}}\mathcal{C}_{t,m}
        +
        \lambda_{\mathrm{prog}}\mathcal{P}_{t,m}
    \right),
    \label{eq:action_selection}
\end{equation}
where $\lambda_{\mathrm{cov}}$ and $\lambda_{\mathrm{prog}}$ balance the two exploration objectives. 

After selection, \textsc{Valerant} restores the pre-action state and commits only $\mathbf{A}^{(m^\star)}_t$. The selected action is replayed from the restored state, while all unsuccessful branches are discarded without modifying the persistent world state. The selected observations and reconstructed geometry are then incorporated into the next visual context and global map:
\begin{equation}
    \begin{aligned}
        \mathbf{O}^{\mathrm{c}}_{t+1}
        &=
        \operatorname{Update}\!\left(
            \mathbf{O}^{\mathrm{c}}_t,
            \mathbf{O}^{\mathrm{f}}_{t,m^\star}
        \right),\\
        \mathcal{M}_{t+1}
        &=
        \operatorname{Fuse}\!\left(
            \mathcal{M}_t,
            \hat{\mathcal{M}}_{t,m^\star}
        \right).
    \end{aligned}
    \label{eq:commit_update}
\end{equation}
This branch--reconstruct--evaluate--commit loop is repeated until the exploration budget is exhausted, progressively consolidating the selected observations into a 3D point-cloud game map. 

\subsection{Robust Exploration under Geometric Drift and Visual Hallucinations}
Generated rollouts may contain visual hallucinations, while monocular SLAM can accumulate pose and scale drift, leading to unreliable geometry and unsafe exploration decisions. We define robust exploration as the ability to maintain geometrically consistent mapping and collision-aware progress under these errors, while recovering from locally trapped states.
\label{sec:robust_exploration}

\subsubsection{Robust Floor Estimation}
\label{sec:robust_floor}

Reliable collision reasoning requires a stable estimate of the floor. In particular, reconstructed points must be distinguished as traversable ground, body-level obstacles, or overhead structures according to their height above the local supporting surface. A floor reference calibrated only once is unreliable under monocular SLAM, whose estimated scale and vertical position may drift over time. As a result, an otherwise valid ground surface can gradually shift into the obstacle height range, producing false collision detections even in open space.

To address this issue, we introduce a robust floor estimator that replaces the fixed global reference with a locally adaptive one. For each SLAM keyframe, we identify candidate support points near the lower envelope of its local reconstruction and fit a dominant plane. The estimate is accepted only when the plane has sufficient geometric support and its normal is consistent with the current vertical direction. When the floor cannot be reliably observed, such as when the camera faces a nearby wall, we propagate the most recent reliable camera-to-floor relation instead of reverting to the initial global calibration. This allows the floor estimate to follow gradual SLAM drift while preventing non-ground surfaces from being adopted as the new reference.

Let $\mathbf{q}_i$ be a point on the estimated floor plane of keyframe $i$, $\mathbf{n}_i$ its unit normal oriented toward the camera center $\mathbf{c}_i$, and $\mathbf{p}$ a reconstructed 3D point. We define its normalized floor-relative height as
\begin{equation}
    \bar{h}_i(\mathbf{p})
    =
    \frac{
        \mathbf{n}_i^{\top}(\mathbf{p}-\mathbf{q}_i)
    }{
        \mathbf{n}_i^{\top}(\mathbf{c}_i-\mathbf{q}_i)
    }.
    \label{eq:robust_floor}
\end{equation}
Under this normalization, the local floor has height zero and the camera center has height one. A common translation or scale change in the local reconstruction therefore affects the floor, camera, and scene points together and is canceled by the relative representation. All subsequent collision tests operate on $\bar{h}_i$ rather than absolute global height. Consequently, the ground and body-level occupancy bands remain aligned under reconstruction drift, substantially reducing false obstacles caused by inconsistent floor elevations.

\subsubsection{Complementary Collision Detection}
\label{sec:collision_detection}

\paragraph{Point-Cloud Collision.}
Action-conditioned video WMs do not explicitly enforce collision constraints and may generate trajectories that pass through scene geometry. We detect such failures directly from the branch-local geometry reconstructed for each candidate rollout, without relying on a learned occupancy predictor. We treat the candidate map $\hat{\mathcal{M}}_{t,m}$ as a set of reconstructed 3D points and use $\mathbf{p}$ to denote an individual point. Each point is assigned a normalized floor-relative height $\bar{h}(\mathbf{p})$ using the local floor estimate of its source keyframe from \eqref{eq:robust_floor}. Let $\mathcal{S}_{t,m}$ denote the set of trajectory samples used for geometric testing, with $\hat{\mathbf{T}}_{t,m,i}$ representing the pose of sample $i\in\mathcal{S}_{t,m}$ and $\kappa_{t,m,i}$ its associated camera-to-floor distance. We detect a point-cloud collision by testing whether the reconstructed geometry intersects a normalized cylindrical body volume along the sampled trajectory:
\begin{equation}
    \mathcal{B}^{\mathrm{pc}}_{t,m}
    =
    \mathbb{I}\!\left[
        \max_{i\in\mathcal{S}_{t,m}}
        \sum_{\mathbf{p}\in\hat{\mathcal{M}}_{t,m}}
        \mathbb{I}\!\left(
            \frac{
                \operatorname{dist}_{\perp}
                (\mathbf{p},\hat{\mathbf{T}}_{t,m,i})
            }{
                \kappa_{t,m,i}
            }
            \leq \rho
            \;\land\;
            \eta_{\mathrm{low}}
            \leq
            \bar{h}(\mathbf{p})
            \leq
            \eta_{\mathrm{high}}
        \right)
        \geq N_{\mathrm{occ}}
    \right],
    \label{eq:pointcloud_collision}
\end{equation}
where $\operatorname{dist}_{\perp}$ measures the horizontal distance from a point to the agent's vertical body axis, $\rho$ is the normalized body radius, and $[\eta_{\mathrm{low}},\eta_{\mathrm{high}}]$ specifies the body-level height range. The threshold $N_{\mathrm{occ}}$ suppresses isolated reconstruction noise, while the height restriction excludes traversable ground and overhead structures. Rotation-only candidates are exempt because they introduce no additional translational penetration.

\paragraph{Depth-Based Collision.}
Point-cloud collision detection may become unreliable for textureless or reflective walls, whose correspondences can receive low confidence and be removed during SLAM reconstruction. We therefore introduce a complementary detector that operates directly on the unfiltered depth estimates of all rollout frames, including those not selected as SLAM keyframes. Let $\boldsymbol{\delta}_{t,m,\ell}$ denote the raw depth map at rollout frame $\ell$, $\Omega_{\mathrm{ctr}}$ rectangular region centered in the image, and $\kappa_{t,m,\ell}$ the camera-to-floor distance associated with that frame. A penetration event is identified when a frontal surface appears within a normalized near-field distance but the estimated trajectory nevertheless continues to advance:
\begin{equation}
    \mathcal{B}^{\mathrm{dep}}_{t,m}
    =
    \mathbb{I}\!\left[
        \exists\,\ell:
        \frac{
            \operatorname{Avg}_{\mathbf{u}\in\Omega_{\mathrm{ctr}}}
            \boldsymbol{\delta}_{t,m,\ell}(\mathbf{u})
        }{
            \kappa_{t,m,\ell}
        }
        \leq d_{\mathrm{near}}
        \;\land\;
        \frac{
            \operatorname{Fwd}
            \!\left(
                \hat{\mathbf{T}}_{t,m};
                \ell,L
            \right)
        }{
            \kappa_{t,m,\ell}
        }
        \geq s_{\mathrm{min}}(\ell)
    \right],
    \label{eq:depth_collision}
\end{equation}
where $\operatorname{Fwd}(\hat{\mathbf{T}}_{t,m};\ell,L)$ measures the forward displacement along the estimated trajectory over the video interval from frame $\ell$ to frame $L$, and $s_{\mathrm{min}}(\ell)$ accounts for the remaining rollout horizon. The central average provides a robust signature of a nearby frontal surface, while the subsequent motion distinguishes penetration from merely observing a close wall. The collision indicator $\mathcal{B}_{t,m}$ in \eqref{eq:action_selection} is activated whenever either $\mathcal{B}^{\mathrm{pc}}_{t,m}$ or $\mathcal{B}^{\mathrm{dep}}_{t,m}$ is positive.

\subsubsection{Dead-Pocket Rewind}
\label{sec:dead_pocket_rewind}

Receding-horizon exploration can lead the agent into a dead pocket in which every translational candidate either violates the collision-free constraint or provides negligible forward progress. Continuing local action selection in this state produces repeated rotations or stagnation rather than revealing new space. We address this failure through a checkpoint-based rewind mechanism that augments local exploration with backtracking. Along the committed trajectory, \textsc{Valerant} maintains synchronized checkpoints of the WM and SLAM states. When insufficient progress persists, the system selects an earlier checkpoint $r<t$ preceding the stalled segment and restores the visual context, camera pose, and reconstructed map to that state. To prevent the agent from immediately re-entering the same pocket, the spatial cells traversed along the discarded segment are retained in a failed-region memory. Let $\boldsymbol{\Phi}_t$ denote the joint WM--SLAM checkpoint at step $t$, let $\mathcal{F}_t$ denote the accumulated set of rejected cells, and let $\operatorname{cell}(\mathbf{T}_k)$ map a camera pose to its spatial cell. The rewind operation is summarized as
\begin{equation}
    \boldsymbol{\Phi}_{t^{+}}
    \leftarrow
    \boldsymbol{\Phi}_{r},
    \qquad
    \mathcal{F}_{t^{+}}
    \leftarrow
    \mathcal{F}_{t}
    \cup
    \left\{
        \operatorname{cell}(\mathbf{T}_k)
        \,\middle|\,
        r<k\leq t
    \right\},
    \label{eq:dead_pocket_rewind}
\end{equation}
where $t^{+}$ denotes the state immediately after rewinding. Subsequent candidate trajectories that enter $\mathcal{F}_{t^{+}}$ are treated as inadmissible, encouraging exploration from the restored anchor toward an alternative route. Because the WM and SLAM states are restored jointly, the generated visual context and reconstructed geometry remain synchronized after backtracking. If the restored state remains trapped, the procedure can be repeated with an earlier checkpoint, turning the otherwise greedy exploration process into a spatial search with recoverable decisions.
\section{Experiments}
\begin{figure}
    \centering
    \includegraphics[width=0.9\linewidth]{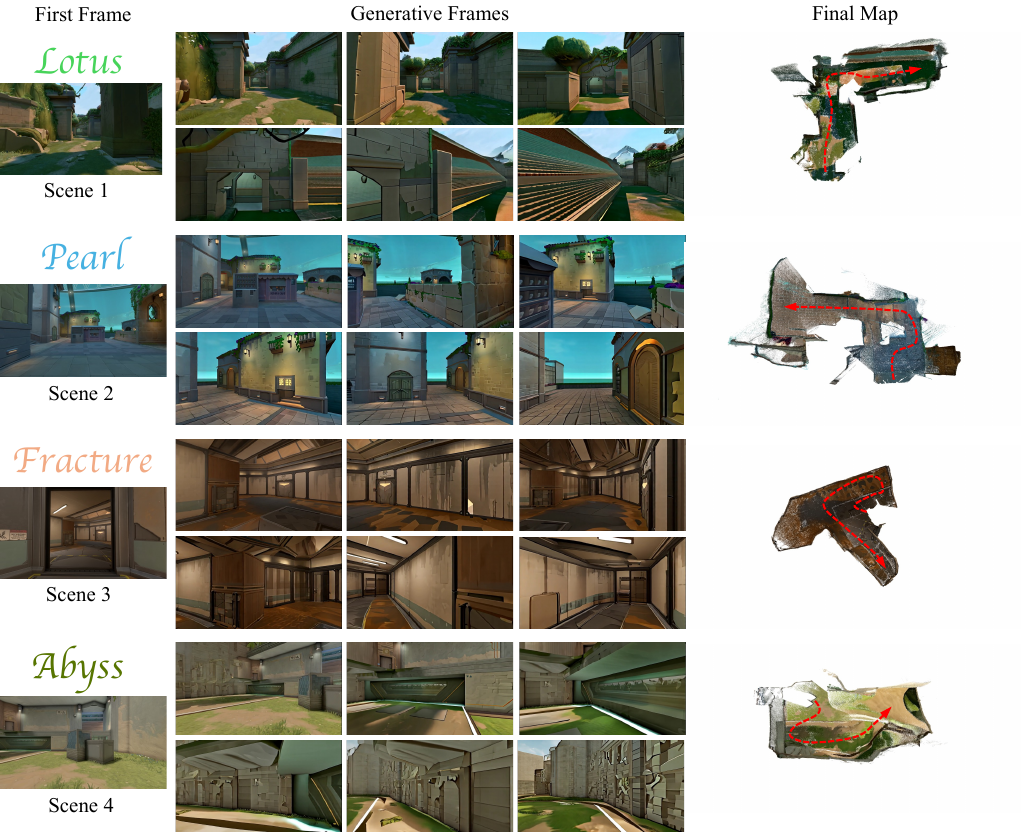}
    \caption{Qualitative Results of \textsc{Valerant} for Single-Image 3D Game-Map Generation. Each pair of rows presents one scene. The leftmost column shows the input first frame, the middle columns show six generated frames sampled along the agent's trajectory, and the rightmost column presents a top-down view of the final reconstructed map with the agent's trajectory overlaid.}
    \label{res_fig1}
    \vspace{-3mm}
\end{figure}
\begin{figure}
    \centering
    \includegraphics[width=0.9\linewidth]{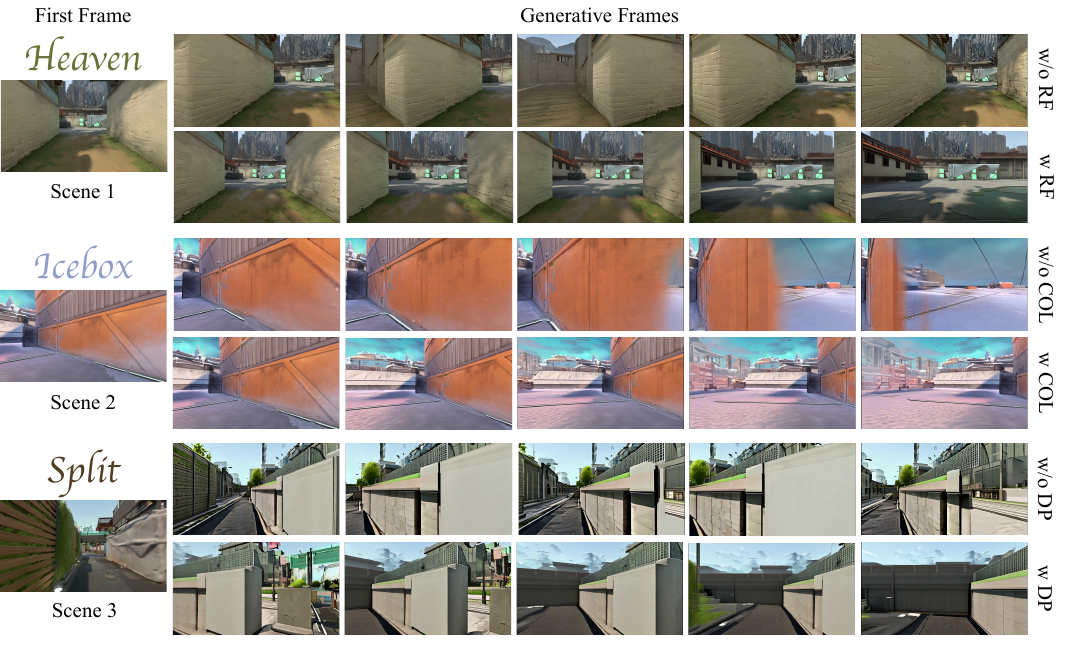}
    \caption{Qualitative ablation of the robust exploration modules. Each pair of rows presents one scene and isolates the effect of a single module by comparing generated frames without (top) and with (bottom) that module. RF, COL, and DP denote robust floor estimation, complementary collision detection, and dead-pocket rewind, respectively.}
    \vspace{-4mm}
    \label{ablation}
\end{figure}
\subsection{Experiment Settings}

\looseness -1
\paragraph{Dataset \& Baseline} All input images used in our experiments are captured directly as screenshots from \textit{VALORANT}. Each screenshot serves as the initial frame and single-view condition. We evaluate all methods using the same set of initial frames. We compare \textsc{Valerant} with HY-World~2.0 \citep{hy2026hy}, which likewise generates a navigable 3D world from a single image but adopts a substantially different technical approach. 

\paragraph{Metric} Our task is inherently open-ended and generative: a single input image can correspond to many plausible 3D environments, and no unique ground-truth map is available for reference-based evaluation. We therefore adopt two perceptual metrics: \textit{Authenticity Score} and \textit{User Preference}. The former uses a VLM as an evaluator to assess the perceptual authenticity of the generated results, while the latter reports the proportion of pairwise comparisons in which human evaluators prefer the output of each method. Detailed results are provided in Appendix~\ref{appendix c}. Unlike training-based benchmarks evaluated on held-out data with paired references, our training-free method operates on arbitrary input images without ground-truth 3D scenes. Since different methods may generate distinct yet plausible content, correspondence-based metrics are not meaningful; we therefore rely on VLM evaluation and user preference.

\paragraph{Implementation Details}
We use Matrix-Game~3.0 \citep{wang2026matrix} as the action-conditioned video WM and MASt3R-SLAM \citep{murai2025mast3r} for visual localization and 3D reconstruction. We optionally apply video2world \citep{hollein2026world} as a post-processing stage to globally refine the reconstructed point cloud. All experiments are conducted on a single NVIDIA H100 GPU. Detailed hyperparameter settings are provided in Appendix \ref{appendix a}.

\vspace{-1mm}
\subsection{Experiment Results}
\vspace{-1mm}
\paragraph{Main Results} Figure~\ref{res_fig1} presents the qualitative results of \textsc{Valerant} across four scenes. Starting from only the first frame, the entire exploration and reconstruction process proceeds automatically, without manually specified trajectories or intermediate intervention. The generated frames sampled along each trajectory show that the agent moves smoothly through corridors, turns, and open regions while avoiding collisions. Meanwhile, visual SLAM continuously integrates the selected observations into a persistent 3D point-cloud map, with the final top-down views demonstrating substantial spatial coverage along the executed trajectories. These results show that \textsc{Valerant} can jointly achieve autonomous collision-free exploration and incremental 3D game-map construction from a single image.

\vspace{-3mm}
\paragraph{Ablation Study} 
Figure~\ref{ablation} qualitatively isolates the contributions of the three robustness components. In Heaven, removing RF destabilizes the estimated ground reference under geometric drift, causing the agent to remain near the surrounding walls and make limited exploration progress; RF restores reliable floor-relative reasoning and enables smooth forward traversal into the open area. In Icebox, the absence of COL allows generated trajectories to penetrate the orange wall, whereas COL rejects geometrically infeasible candidates and keeps exploration within free space. In Split, without DP, the agent becomes trapped in a dead pocket and repeatedly alternates between left and right turns without making spatial progress; DP restores an earlier exploration state and redirects the agent toward an alternative route. These comparisons demonstrate that RF, COL, and DP address complementary failure modes arising from floor-estimation drift, wall penetration, and local exploration traps, respectively.
\vspace{-3mm}
\section{Conclusion}
\vspace{-2mm}
We presented \textsc{Valerant}, a training-free framework that constructs navigable 3D game maps from a single image. By coupling a frozen action-conditioned video world model with visual SLAM and an exploration policy, \textsc{Valerant} evaluates alternative action rollouts and incrementally integrates selected observations into a persistent point-cloud map. Experiments show that robust floor estimation, collision detection, and dead-pocket rewind support reliable autonomous exploration under geometric drift and visual hallucinations. Future work will improve geometric consistency and convert the reconstructed point clouds into richer, editable game assets.

\bibliography{iclr2027_conference}
\bibliographystyle{iclr2027_conference}

\clearpage
\appendix
\section{Implementation Details}
\begin{figure}[]
    \centering
    \includegraphics[width=1\linewidth]{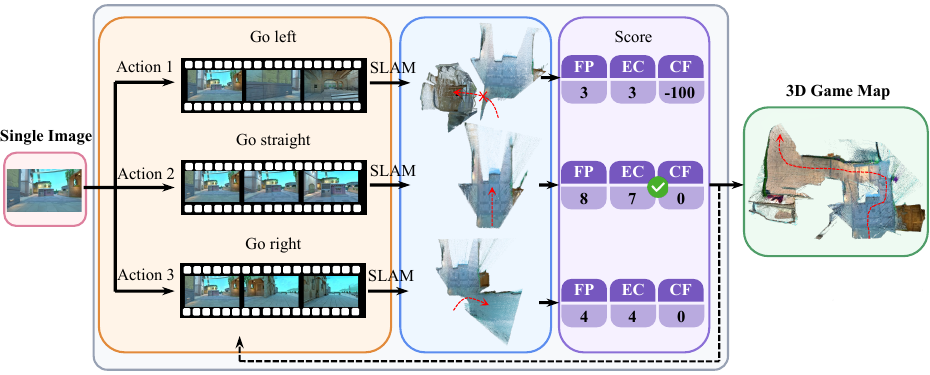}
    \caption{A detailed version of the pipeline, expanding upon Figure~\ref{fig1}.}
    \label{fig2}
\end{figure}
\label{appendix a}

\paragraph{World Model and Reconstruction.}
We instantiate $p_{\theta}$ with the frozen Matrix-Game~3.0 action-conditioned video WM. The visual context length is set to $K=16$ frames. For the single-stage candidates $\mathtt{uw}$, $\mathtt{jq}$, and $\mathtt{lq}$, the prediction horizon is $H=57$ frames. The composite candidates $\mathtt{jw}$ and $\mathtt{lw}$ append a 40-frame forward-motion segment after the initial turn, resulting in $H=97$ frames. Flow-matching inference uses three integration steps with a classifier-free guidance scale of $5.0$. At every exploration step, all five candidates in $\mathcal{A}_t$ are generated from the same committed visual context. We use MASt3R-SLAM to estimate $\hat{\mathbf{T}}_{t,m}$ and reconstruct $\hat{\mathcal{M}}_{t,m}$ for each rollout. When specified, video2world is applied only as a post-processing stage to refine the final point cloud and does not participate in action selection.

\paragraph{Exploration Policy.}
The coverage utility is instantiated as
$\mathcal{C}_{t,m}
=
\mathcal{C}^{\mathrm{new}}_{t,m}
+
0.5\mathcal{C}^{\mathrm{front}}_{t,m}$,
where the two terms measure newly visited cells and proximity to the current exploration frontier, respectively. In ~\eqref{eq:action_selection}, we set
$\lambda_{\mathrm{cov}}=1.0$ and
$\lambda_{\mathrm{prog}}=5.0$.
Both coverage estimation and the failed-region memory $\mathcal{F}_t$ operate on a $128\times128$ spatial grid whose cell width is one eighth of the nominal displacement of a forward action. The exploration budget is limited to 15 committed steps.

\paragraph{Robust Floor Estimation.}
For each SLAM keyframe, a candidate floor plane is accepted only when it is supported by at least 300 reconstructed points and satisfies
$\lvert\mathbf{n}_i^{\top}\mathbf{e}_y\rvert\geq0.7$,
where $\mathbf{e}_y$ denotes the vertical axis. If either condition is not met, the most recent reliable camera-to-floor relation is propagated. All subsequent geometric tests use the normalized height $\bar{h}_i(\mathbf{p})$ from ~\eqref{eq:robust_floor}.

\paragraph{Collision Detection.}
For point-cloud collision detection, the normalized body radius in ~\eqref{eq:pointcloud_collision} is set to $\rho=0.21$, and the body-level height interval is
$[\eta_{\mathrm{low}},\eta_{\mathrm{high}}]=[0.18,1.19]$.
We evaluate every pose in $\mathcal{S}_{t,m}$ and retain reconstructed points with a confidence score of at least $4.0$. The occupancy threshold is $N_{\mathrm{occ}}=30$ and must be exceeded at two consecutive trajectory samples; a single sample containing at least 300 occupied points is treated as an immediate deep-penetration event. Any detected collision activates the hard constraint $\mathcal{B}_{t,m}=1$.

For depth-based collision detection, $\Omega_{\mathrm{ctr}}$ covers the central one-third of the image along both height and width, and every rollout frame is evaluated. The near-field cutoff is $0.45$ in reconstructed depth units, corresponding to $d_{\mathrm{near}}=0.455$ in the normalized form of ~\eqref{eq:depth_collision}. We set
$s_{\min}(\ell)
=
0.085\,d_{\mathrm{fwd}}(L-\ell)/L$,
where $d_{\mathrm{fwd}}$ is the normalized displacement of a nominal forward action. This decay accounts for the amount of rollout remaining after frame $\ell$.

\looseness -1
\paragraph{Dead-Pocket Rewind.}
A dead pocket is declared when the policy selects rotation-only actions for three consecutive steps without translational progress. \textsc{Valerant} retains the three most recent synchronized WM--SLAM checkpoints $\boldsymbol{\Phi}_t$ and permits at most two rewind operations during one exploration episode. After rewinding, cells belonging to the discarded trajectory are added to $\mathcal{F}_t$ according to ~\eqref{eq:dead_pocket_rewind}, preventing the policy from immediately entering the same failed region.

All experiments are conducted at inference time on a single NVIDIA H100 GPU, without additional training or fine-tuning of the WM or SLAM modules.
\section{Metric Details}
\label{appendix b}

\paragraph{VLM Evaluation Prompt.}
\begin{quote}
\itshape

You are a professional 3D scene evaluator with strong visual judgment. Your task is to compare the quality of 3D Gaussian Splatting (3DGS) scenes generated by two different methods.

For each evaluation case, I will provide:
\begin{enumerate}
    \item A first frame shared by both methods, which serves as the common generation input and defines the initial appearance and visible content of the scene.
    \item Five rendered images of a 3DGS scene generated by Method A.
    \item Five rendered images of a 3DGS scene generated by Method B.
\end{enumerate}

Within each method, the five rendered images are presented in their camera-path order and depict different viewpoints of the same generated scene. The method labels are anonymized and arbitrary.

You must compare the two generated scenes and make a single-choice decision: select the better result between Method A and Method B. Please base your judgment on the following criteria:

\begin{enumerate}
    \item \textbf{Overall visual quality:}
    Evaluate the immediate visual impression of each result at first glance. Consider whether the scene appears visually appealing, coherent, complete, and free from conspicuous defects. Judge each result as a whole rather than focusing on a single isolated image or detail.

    \item \textbf{3D scene consistency:}
    Assess whether the geometry, spatial layout, surfaces, and scene structures remain consistent across the five rendered viewpoints. Scene elements visible in the shared first frame should be preserved plausibly when observed from subsequent viewpoints. Look for distorted geometry, duplicated or missing structures, floating elements, broken surfaces, and viewpoint-dependent artifacts.

    \item \textbf{Cross-view and trajectory coherence:}
    Determine whether the shared first frame and the five subsequent renders appear to form a natural sequence captured from one persistent scene along a plausible camera path. Shared structures should remain recognizable, while perspective, visibility, and scene content should change smoothly as the camera moves. Penalize abrupt changes that make the renders appear independently generated or suggest that they belong to different scenes.

    \item \textbf{Game-scene realism and rendering quality:}
    Evaluate whether the result resembles a high-quality scene rendered from a 3D game environment. Consider the quality of textures, materials, lighting, geometry, details, and overall rendering, as well as the absence of blur, holes, floaters, or other visible artifacts. Photographic realism is not required; the scene should instead look visually credible and internally coherent as a game environment.
\end{enumerate}

After considering all four criteria, choose the single better result between the two methods. Use the shared first frame only as a common reference and evaluate the five rendered images of each method jointly. Do not select a method based on only one view. Your decision should be based solely on the provided visual evidence, without making assumptions about the identities or technical implementations of the methods. Ties are not allowed.

Please output only the final choice in the following format:

\begin{center}
    \texttt{Best Method: [A/B]}
\end{center}

\end{quote}
For the user study, participants compared the two methods following the same instructions and four evaluation criteria used in the VLM-based assessment. Their preferences were collected through a questionnaire, and participation entailed no foreseeable risk.
\begin{figure}
    \centering
    \includegraphics[width=1\linewidth]{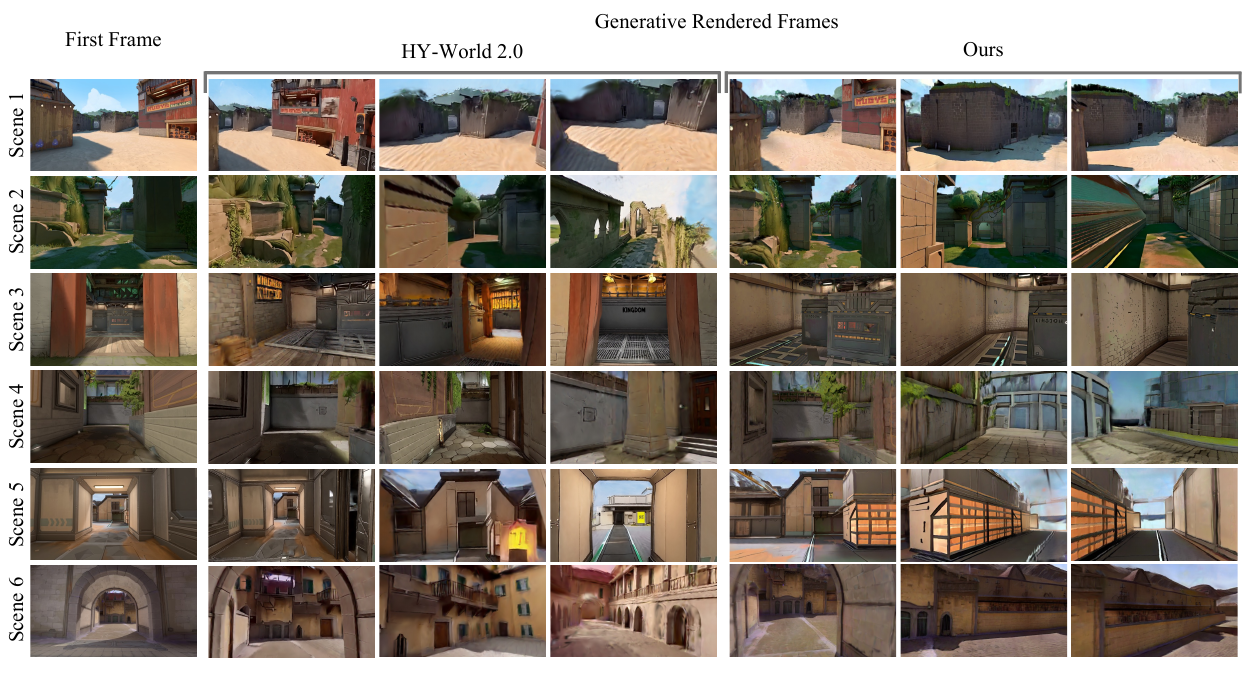}
    \looseness -1
    \caption{\textbf{Qualitative comparison.} Starting from the same input frame, both methods construct a 3DGS scene, from which three views sampled along a camera trajectory are rendered. Compared with HY-World~2.0, \textsc{Valerant} produces more coherent geometry and appearance across viewpoints, resulting in a more consistent and natural representation of the generated scene.}
    \label{fig_appen}
\end{figure}
\section{Baseline Comparison}
\label{appendix c}

\looseness -1
To the best of our knowledge, HY-World~2.0 \citep{hy2026hy} is the only publicly available baseline that shares our objective of constructing an explorable 3D scene from a single image. Given an input image, HY-World~2.0 first employs HY-Pano~2.0 to synthesize a complete $360^{\circ}$ panorama. WorldNav then parses the panorama into preliminary geometry, semantic landmarks, and navigable regions, from which it plans camera trajectories intended to maximize scene coverage. Following these trajectories, WorldStereo~2.0 generates camera-controlled keyframes to expand regions that are not sufficiently observed in the initial panorama. Finally, WorldMirror~2.0 reconstructs the generated multi-view observations, after which the resulting geometry is composed and optimized into a 3DGS scene.

HY-World~2.0 produces sparse keyframe observations rather than directly returning a dense video sequence comparable to the action-conditioned rollouts generated by \textsc{Valerant}. For a fair comparison in a common output space, we therefore represent and render the results of both methods as 3DGS scenes. Consequently, the visual quality of both sets of rendered images is constrained by the reconstruction and optimization of their 3DGS representations and may be lower than that of directly generated video frames. Figure~\ref{fig_appen} presents a qualitative comparison between the two methods under this unified setting.

As shown in Figure~\ref{fig_appen}, HY-World~2.0 exhibits weaker cross-view consistency and less coherent scene structure. We attribute this difference primarily to its sequential generation process. Its visual content is synthesized in two successive phases: an initial panorama is first generated from the input image, after which a separate world-expansion stage synthesizes additional observations for regions that are missing or insufficiently covered by the panorama. Although HY-World~2.0 introduces geometric and stereo memory to improve consistency, discrepancies introduced during panorama generation can still be propagated or amplified when the missing regions are subsequently completed. The expanded views may therefore disagree with the initial panorama or with one another, producing appearance drift and geometric discontinuities in the final 3DGS reconstruction. In contrast, \textsc{Valerant} generates action-conditioned video rollouts from the currently committed context and incrementally integrates selected observations, which better preserves local continuity along the exploration trajectory.

The quantitative results in Table~\ref{tab:quantitative_comparison} further support these observations. \textsc{Valerant} achieves an AS of $80.00\%$, compared with $20.00\%$ for HY-World~2.0, indicating that the VLM evaluator more frequently favors its overall visual quality, 3D consistency, trajectory coherence, and game-scene realism. Human evaluation exhibits a similar trend: \textsc{Valerant} receives $83.33\%$ of user preferences, while HY-World~2.0 receives $16.67\%$. The close agreement between the VLM-based assessment and human preferences suggests that the qualitative advantages of \textsc{Valerant} are consistently perceived under both evaluation protocols.

\begin{table}[]
    \centering
    \caption{Quantitative comparison. AS denotes aesthetic score, and UP denotes user preference.}
    \label{tab:quantitative_comparison}
    \setlength{\tabcolsep}{4pt}
    \begin{tabular}{lcc}
        \toprule
        Method
        & AS (\%)$\uparrow$
        & UP (\%)$\uparrow$ \\
        \midrule
        HY-World 2.0 & 20.00 & 16.67 \\
        \textbf{\textsc{Valerant} (Ours)} & 80.00 & 83.33 \\
        \bottomrule
    \end{tabular}
\end{table}

\end{document}